\documentclass{sig-alternate}
\usepackage{graphicx}
\usepackage{amssymb}
\usepackage{times}
\usepackage{verbatim}
\usepackage{amsmath}
\usepackage{algorithmic,algorithm}
\DeclareGraphicsExtensions{.eps}
\graphicspath{{figures/}}

\begin{document}
\conferenceinfo{GECCO'08,} {July 12--16, 2008, Atlanta, Georgia,
USA.}
\CopyrightYear{2008}
\crdata{978-1-60558-131-6/08/07}

\title{A Study of NK Landscapes' Basins and \\ Local Optima
Networks} \numberofauthors{4}
\author{
\alignauthor
Gabriela Ochoa\\
       \affaddr{Automated Scheduling, Optimisation and Planning}\\
       \affaddr{School of Computer Science}\\
       \affaddr{University of Nottingham, UK}\\
       \email{gxo@cs.nott.ac.uk}
\alignauthor
Marco Tomassini\\
       \affaddr{Information Systems Department}\\
       \affaddr{University of Lausanne}\\
       \affaddr{Lausanne, Switzerland}\\
       \email{Marco.Tomassini@unil.ch}
\alignauthor Seb{\'a}stien  V{\'e}rel\\
       \affaddr{Laboratoire I3S}\\
       \affaddr{CNRS-University of Nice}\\
       \affaddr{Sophia Antipolis, France}\\
       \email{verel@i3s.unice.fr}
\and  
\alignauthor Christian Darabos\\
       \affaddr{Information Systems Department}\\
       \affaddr{University of Lausanne}\\
       \affaddr{Lausanne, Switzerland}\\
       \email{Christian.Darabos@unil.ch}
}
\date{}

\maketitle
\begin{abstract}
We propose a network characterization of combinatorial fitness
landscapes by adapting the notion of {\em inherent networks}
proposed for energy surfaces~\cite{doye02}. We use the well-known
family of $NK$ landscapes as an example. In our case the inherent
network is the graph where the vertices are all the local maxima and
edges mean basin adjacency between two maxima. We exhaustively
extract such networks on representative small $NK$ landscape
instances, and show that they are `small-worlds'. However, the
maxima graphs are not random, since their clustering coefficients
are much larger than those of corresponding random graphs.
Furthermore, the degree distributions are close to exponential
instead of Poissonian. We also describe the nature of the basins of
attraction and their relationship with the local maxima network.
\end{abstract}


\category{I.2.8}{Artificial Intelligence}{Problem Solving, Control
Methods, and Search}[Heuristic methods] \category{G.2.2}{Discrete
Mathematics}{Graph Theory}[Network problems]

\terms{Algorithms, Measurement, Performance}

\keywords{Landscape Analysis, Network Analysis, Complex Networks,
Local Optima, $NK$ Landscapes}

\section{Introduction}

A fitness landscape of a combinatorial problem can be seen as a
graph whose vertices are the possible configurations. If two
configurations can be transformed into each other by a suitable
operator move, then we can trace an edge between them. The resulting
graph, with an indication of the fitness at each vertex, is a
representation of the given problem fitness landscape.
Doye~\cite{doye02,doye05} has recently introduced a useful
simplification of the fitness landscape graph for the energy
landscapes of atomic clusters. The idea consists in taking as
vertices of the graph not all the possible configurations, but only
those that correspond to energy minima. For atomic clusters these
are well-known, at least for relatively small assemblages. Two
minima are considered connected, and thus an edge is traced between
them, if the energy barrier separating them is sufficiently low. In
this case there is a transition state, meaning that the system can
jump from one minimum to the other by thermal fluctuations going
through a saddle point in the energy hyper-surface. The values of
these activation energies are mostly known experimentally or can be
determined by simulation. In this way, a network can be built which
is called the ``inherent structure'' or ``inherent network''
in~\cite{doye02}. We use a modification of this idea for studying
the well-known $NK$ combinatorial landscapes. In our case, a vertex
of the graph is a local maximum, and there is an edge between two
maxima if they lay on adjacent basins.

In the context of meta-heuristics, it is important to identify the
features of landscapes that would influence the effectiveness of
heuristic search. Such knowledge may be helpful for both predicting
the performance and improving the design of meta-heuristics. Among
the features of landscapes known to have a strong influence on
heuristic search, is the number and distribution of local optima in
the search space. An interesting property of combinatorial
landscapes, which has been observed in many different studies, is
that on average, local optima are very much closer to the global
optimum than are randomly chosen points, and closer to each other
than random points would be. In other words, the local optima are
not randomly distributed, rather they tend to be clustered in a
"central massif" (or ``big valley'' if we are minimising). This
globally convex landscape structure has been observed in the $NK$
family of landscapes \cite{kauffman93}, and in many combinatorial
optimisation problems, such as the traveling salesman problem
\cite{boese94}, graph bipartitioning \cite{merz98}, and flowshop
scheduling \cite{reeves99}.

In this study we seek to provide fundamental new insights into the
structural organization of the local optima in combinatorial
landscapes,  particularly into the connectivity and characteristics
of their basins of attraction, using $NK$ landscapes as a case
study. To achieve this, we first map the landscape onto a network,
and then analyze the topology of this network for a number of small
$NK$ landscape instances for which complete networks can be
obtained. Our analysis is inspired, in particular, by the work of
Doye~\cite{doye02,doye05} on energy landscapes, and in general, by
the field of complex networks~\cite{newman03,watts04,watts98}. The
study of complex networks has already permeated the evolutionary
computation field. Specifically, in the study of scientific
collaborations \cite{cotta06,lutti07}, the structure of a population
in cellular evolutionary algorithms
\cite{giacobini06,giacobini05,payne07}, and the evolution of
networks of cellular automata \cite{tomassini04}. However, our study
is the first attempt, to our knowledge, of using network analysis
techniques in connection with the study of fitness landscapes and
problem difficulty in combinatorial optimization.

The next section introduces the study of complex networks, and
describes the main features of {\em small-world} and {\em
scale-free} networks. Section \ref{landscapes} describes how
landscapes are mapped onto networks, and includes the relevant
definitions and algorithms. The empirical network analysis of our
selected $NK$ landscape instances is presented in Section
\ref{analysis}, whilst Section \ref{conclusions} gives our
conclusions and ideas for future work.

\section{Complex Networks}
\label{networks}

The recent interest in the study of networks and networked systems
was influenced by the seminal paper by Watts and Strogatz
\cite{watts98}, who showed that many real-world networks are neither
completely ordered nor completely random, but rather exhibit
important properties of both. Some of these network properties can
be quantified by simple statistics such as the clustering
coefficient $C$, which is a measure of local density, and the
average shortest path length $l$, which is a global measure of
separation. It has been shown in recent years that many social,
biological, and man-made system show what has been called a {\em
small-world} topology \cite{watts98}, in which nodes are highly
clustered yet the path length between them is small.

A second important aspect in the study of networks has been the
realization that in many real-world networks, the distribution of
the number of neighbours (the degree distribution) is typically
right-skewed with a "heavy tail", meaning that most of the nodes
have less-than-average degree whilst a small fractions of hubs have a large
number of connections. These qualitative description can be described
mathematically by a power-law \cite{barabasi99}, which has the
asymptotic form $p(k) \sim k^{-\alpha}$. This means that the
probability of a randomly chosen point having a degree $k$ decays
like a power of $k$, where the exponent $\alpha$ (typically in the
range $[2,3]$) determines the rate of decay. A distinguishing
feature of power-law distributions is that when plotted on a double
logarithmic scale, a power-law appears as a straight line with
negative slope $\alpha$. This behavior contrasts with a normal
distribution which would curve sharply on a log-log plot, such that
the probability of a node having a degree greater than a certain
"cutoff" value is nearly zero. The mean would then trivially represent a
characteristic scale for the network degree distribution. Since
networks with power-low degree distribution lack any such cutoff
value, at least in theory, they are often called {\em scale-free} networks
\cite{watts04}. Examples of such scale-free networks are the
world-wide-web, the internet, scientific collaboration and citation
networks, and biochemical networks.

\section{Landscapes as Networks}
\label{landscapes}

To model a physical energy landscape as a network, Doye
\cite{doye05} needed to decide first on a definition both of a state
of the system and how two states were connected. The states and
their connections will then provide the nodes and edges of the
network. For systems with continuous degrees of freedom, the author
achieved this through the `inherent structure' mapping
\cite{stillinger95}. In this mapping each point in configuration
space is associated with the minimum (or `inherent structure')
reached by following a steepest-descent path from that point. This
mapping divides configuration into basins of attraction surrounding
each minimum on the energy landscape.

\begin{figure}[ht]
\begin{center}
    \begin{minipage}[t]{5.2cm}
       \includegraphics[width=5cm]{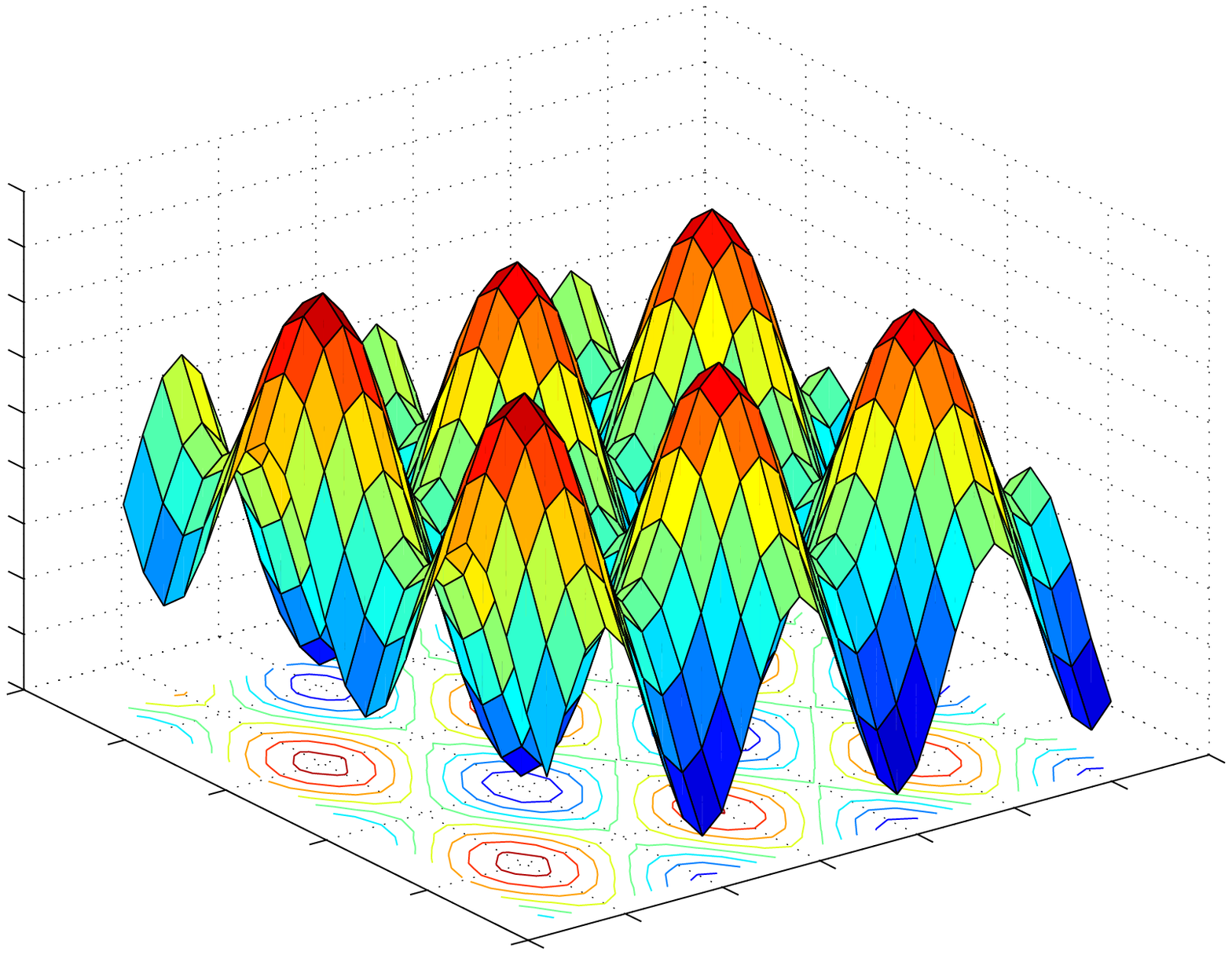}
    \end{minipage}
    \begin{minipage}[t]{3.0cm}
        \includegraphics[width=2.5cm]{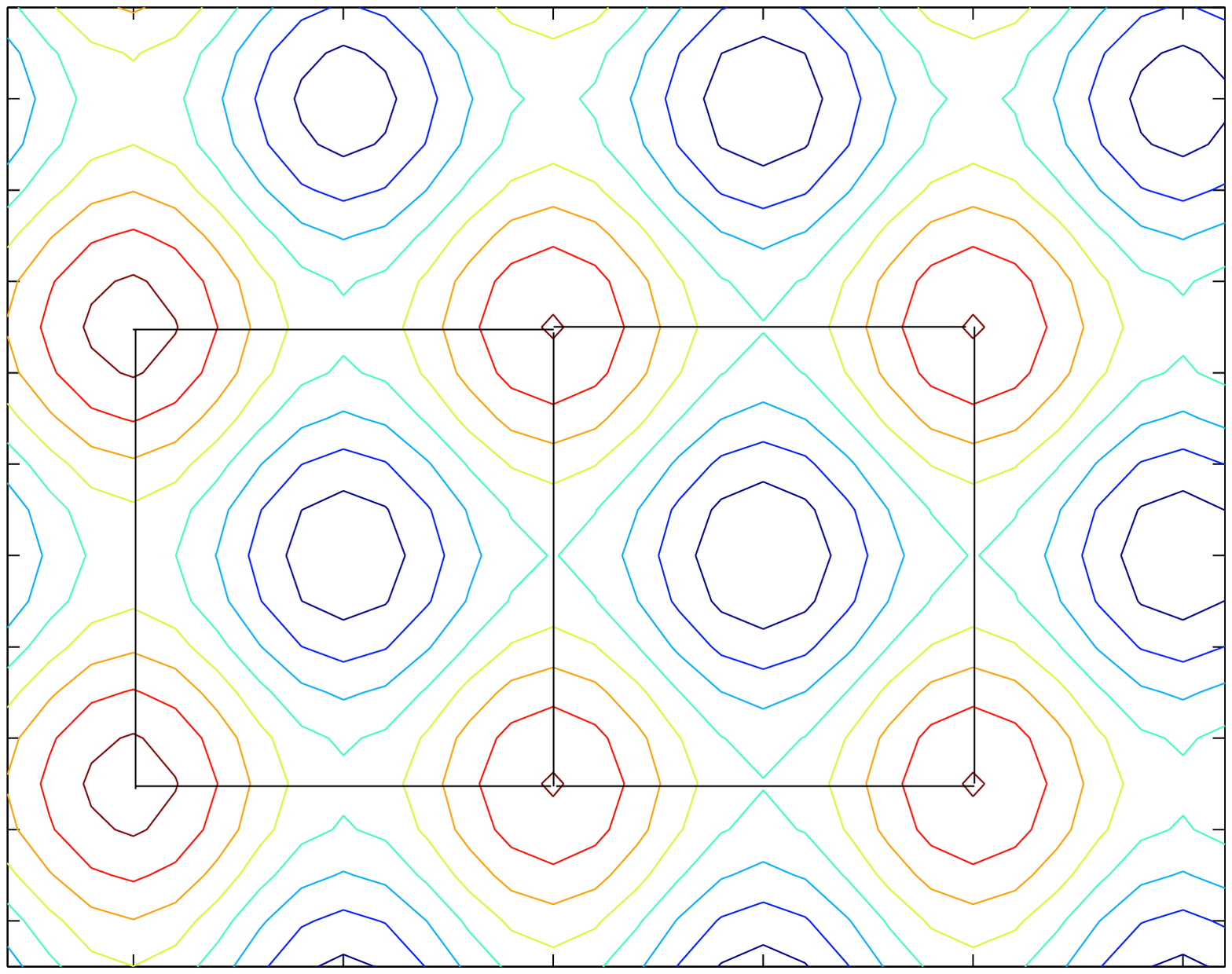}
    \end{minipage}
\caption{A model of a 2D landscape (left), and a contour plot of the
local optima partition of the configuration space into basins of
attraction surrounding maxima and minima (right). A simple regular
network of six local maxima can be observed. \label{fig:inherentn}}
\end{center}
\end{figure}

We use a modification of this idea for the $NK$ family of binary
landscapes, which indeed can be applied to any combinatorial
landscape. In our case, the vertexes of the graph are the local
maxima of the landscape, obtained exhaustively by running a
best-improvement local search algorithm (see Algorithm 1) from every
configuration of the search space. The edges in the network connect
local optima of adjacent basins of attraction. An illustration for a
model 2D landscape can be seen in Figure \ref{fig:inherentn}, which
is inspired by a similar figure appearing in \cite{doye02,doye05}.
Here, we illustrate a network of local maxima (instead of local
minima).  A more formal definition of our inherent networks is given
in Section \ref{defs}. As it was the case in the study on physical
energy landscapes \cite{doye05}, we do not consider multiple edges,
or weights in the edges. This may be a factor to consider in future
work.

Note that while a physical energy landscape is formally a continuous
landscape, ours are strictly combinatorial, i.e. discrete and
finite. Moreover, the energy landscape of a stable atomic cluster,
crystal or molecule is  relatively smooth and easy to search
and has been called a ``funnel'' landscape ~\cite{doye02}.
In contrast, in $NK$ landscapes one can continuously vary the
intrinsic landscape difficulty by changing the value of $K$. As a
result, we shall see that $NK$ landscapes show a number of different
behaviors depending on $K$ for a given $N$, and these different
behaviors are reflected on their inherent networks. Indeed, $NK$
landscapes can be seen as analogous to those of
spin-glasses~\cite{kauffman93,Stein}. In contrast to atomic cluster energy
landscapes, spin glass landscapes may show frustration, i.e.
configurations that must respect conflicting constraints, and
solving for the ground state of the system that is, the minimum
energy configuration is an NP-hard problem. Similar consequences are
caused by the introduction of epistatic interactions through the
increase of the $K$ value in $NK$ landscapes.

Below we present the relevant formal definitions and algorithms to
obtain our combinatorial analogous of an energy landscape inherent
network.

\subsection{Definitions and Algorithms}
\label{defs}

\textbf{Definition : }Fitness landscape.\\
A landscape is a triplet $(S, V, f)$ where $S$ is a set of potential
solutions i.e. a search space, $V : S \longrightarrow 2^S$, a
neighborhood structure, is a function that assigns to every $s \in
S$ a set of neighbours $V(s)$, and $f : S \longrightarrow R$ is a
fitness function that can be pictured as the \textit{height} of the
corresponding potential solutions.

In our study, the search space is composed by binary strings of
length $N$, therefore its size is $2^N$. The neighborhood is defined
by the minimum possible move on a binary search space, that is, the
1-move or bit-flip operation. In consequence, for any given string
$s$ of length $N$, the neighborhood size is $|V(s)| = N$. Notice
that in $NK$ landscapes, two neighboring solutions never have the
same fitness value. Therefore, neutrality is not present. Landscapes
with neutrality will be considered in future work.

\textbf{Definition:} Local Optimum.\\
A local optimum is a solution $s^{*}$ such that $\forall  s \in
V(s^{*})$, $f(s) < f(s^{*})$.

The $LocalSearch$ algorithm to determine the local optima and
therefore define the basins of attraction, is given below:

\begin{algorithm}
\caption{{\em LocalSearch}} \label{algoHC}
\begin{algorithmic}
\STATE Choose initial solution $s \in \cal S$ \REPEAT
    \STATE choose $s^{'} \in V(s)$ such that $f(s^{'}) = max_{x \in {\cal V}(s)}\ f(x)$
        \IF{$f(s) < f(s^{'})$}
            \STATE $s \leftarrow s^{'}$
    \ENDIF
\UNTIL{$s$ is a  Local optimum}
\end{algorithmic}
\end{algorithm}

The $LocalSearch$ algorithm defines a mapping from the search space
$S$ to the set of locally optimal solutions $S^*$. We therefore
define a basin of attraction as follows:

\textbf{Definition : } Basin of attraction.\\
The basin of attraction of a local optimum $i$ is the set $b_i = \{ s
\in S ~|~ LocalSearch(s) = i \}$.
The size of the basin of attraction of a local optima $i$ is the
cardinality of $b_i$.

We then define the inherent network, or network of local
optima as:

\textbf{Definition :} Local optima network.\\
The local optima network $G=(S^*,E)$ is the graph where the nodes are
the local optima, and there is an edge $e_{ij} \in E$ between two local optima
 $i$ and $j$ if there is at least a pair of direct neighbors (1-bit apart) $s_i$ and $s_j$, such that $s_i \in
b_i$ and $s_j \in b_j$. That is, if there exists a pair of direct
neighbors solutions $s_i$ and $s_j$, one in each basin ($b_i$ and
$b_j$)

\section{Empirical Network Analysis}
\label{analysis}

\subsection{Experimental Setting}

The $NK$ family of landscapes \cite{kauffman93} is a
problem-independent model for constructing multimodal landscapes
that can gradually be tuned from smooth to rugged. In the model, $N$
refers to the number of (binary) genes in the genotype (i.e. the
string length) and $K$ to the number of genes that influence a
particular gene. By increasing the value of $K$ from 0 to $N-1$,
$NK$ landscapes can be tuned from smooth to rugged. The $k$
variables that form the context of the fitness contribution of gene
$s_i$ can be chosen according to different models. The two most
widely studied models are the {\em random neighborhood} model, where
the $k$  variables are chosen randomly according to a uniform
distribution among the $n-1$ variables other than $s_i$, and the
{\em adjacent neighborhood} model, in which the $k$ variables that
are closest to $s_i$ in a total ordering $s_1, s_2, \ldots, s_n$
(using periodic boundaries). No significant differences between the
two models were found in \cite{kauffman93} in terms of  global
properties of the respective families of landscapes, such as mean
number of local optima or autocorrelation length. Therefore, we
explore here the adjacent neighborhood model, leaving the random
model for future analysis.

In order to avoid sampling problems that could bias the results, we
used the largest values of $N$ that can still be analyzed
exhaustively with reasonable computational resources. We thus
extracted the local optima networks of landscape instances with $N =
{16, 18}$, and $K = {2,4,6,..., N-2,N-1}$. For each pair of $N$ and
$K$ values, 30 instances were explored. Therefore, the networks
statistics reported below represent the average behaviour of 30
independent instances.

\subsection{General Network Statistics}

Table \ref{tab:statistics} reports the average of the network
properties  measured on $NK$ landscapes for $N= 16, 18$  and all
even $K$ values; $K=N-1$ is also given. Values are averages over 30
randomly generated landscapes. $\bar n_v$ and $\bar n_e$ are,
respectively, the mean number of vertices and the mean number of
edges of the graph for a given $K$ rounded to the next integer.
$\bar C$ is the average of the mean clustering
coefficients\footnote{The clustering coefficient $C_i$ of a node $i$
is defined as $C_i=2E_i/k_i(k_i-1)$, where $E_i$ is the number of
edges in  the neighborhood of $i$. Thus $C_i$ measures the amount of
``cliquishness'' of the neighborhood of node $i$ and it
characterizes the extent to which nodes adjacent to node $i$ are
connected to each other. The clustering coefficient of the graph is
simply the average over all nodes: $C = \frac{1}{N} \sum_{i=1}^{N}
C_i$~\cite{newman03}.} over all the generated landscapes. $C_r$ is
the average clustering coefficient of a random graph with the same
number of vertices and mean degree. $\bar z$ is the average of the
mean degrees. $\bar l$ is the average of the mean path lengths over
all landscape instances. The last column contains the average degree
assortativity coefficient $\bar a$, which measures whether nodes
with similar degrees tend to pair up with each other. The
assortativity coefficient is computed according
to~\protect\cite{newman03}.

\begin{table*}[!ht]
\begin{center}
\small \caption{Network properties of $NK$ landscapes  for $N=16,
18$ and all even $K$ values; $K=N-1$ is also given. Values are
averages over 30 randomly generated landscapes, standard deviations
are shown as subscripts. $n_v$ and $n_e$ represent the number of
vertexes and edges (rounded to the next integer), $\bar C$, the mean
clustering coefficient, whilst $C_r$ is the clustering coefficient
of a random graph with the same number of vertexes and mean degree,
which is $C_r \simeq \bar z / \bar n_v$. $\bar z$ represent the mean
degree, $\bar l$ the mean path length , and $\bar a$ the degree
assortativity coefficient.} \label{tab:statistics}
\begin{tabular}{|c|c|c|c|c|c|c|c|}
\hline
\multicolumn{8}{|c|}{$N = 16$} \\
\hline
 $K$ & $\bar n_v$ & $\bar n_e$ & $\bar C$ & $C_r$ & $\bar z$ & $\bar l$ & $\bar a$ \\
\hline
2  &    $33_{15}$ &     $261_{ 166}$ & $0.68_{0.095}$ & $0.507_{0.1536}$  &  $14.55_{3.826}$ & $1.54_{0.182}$  & $-0.0007_{0.00591}$ \\
4  &   $178_{33}$ &   $6,334_{1646}$ & $0.66_{0.036}$ & $0.406_{0.0615}$  &  $70.48_{6.615}$ & $1.60_{0.062}$  & $-0.0162_{0.00467}$ \\
6  &   $460_{29}$ &  $26,414_{2035}$ & $0.55_{0.013}$ & $0.250_{0.0150}$  & $114.76_{3.033}$ & $1.75_{0.016}$  & $-0.0237_{0.00283}$ \\
8  &   $890_{33}$ &  $56,022_{1951}$ & $0.44_{0.008}$ & $0.139_{0.0061}$  & $124.52_{1.800}$ & $1.88_{0.008}$  & $-0.0219_{0.00250}$ \\
10 & $1,470_{34}$ &  $86,446_{1766}$ & $0.36_{0.006}$ & $0.080_{0.0023}$  & $117.62_{1.137}$ & $2.00_{0.009}$  & $-0.0170_{0.00182}$ \\
12 & $2,254_{32}$ & $117,085_{1111}$ & $0.30_{0.003}$ & $0.046_{0.0009}$  & $103.91_{0.695}$ & $2.19_{0.012}$  & $-0.0122_{0.00104}$ \\
14 & $3,264_{29}$ & $146,390_{1025}$ & $0.26_{0.002}$ & $0.027_{0.0003}$  &  $89.70_{0.349}$ & $2.47_{0.009}$  & $-0.0092_{0.00064}$ \\
15 & $3,868_{33}$ & $160,690_{ 829}$ & $0.25_{0.002}$ & $0.021_{0.0003}$  &  $83.09_{0.469}$ & $2.58_{0.007}$  & $-0.0086_{0.00059}$ \\
\hline
\hline
\multicolumn{8}{|c|}{$N = 18$} \\
\hline
2  &     $50_{25}$ &     $478_{ 342}$ & $0.62_{0.106}$ & $0.414_{0.1697}$ &  $17.08_{4.930}$ & $1.66_{0.210}$ & $0.00390_{0.00530}$  \\
4  &    $330_{72}$ &  $17,576_{4898}$ & $0.61_{0.044}$ & $0.332_{0.0573}$ & $105.39_{8.106}$ & $1.67_{0.058}$ & $-0.0168_{0.00495}$ \\
6  &    $994_{73}$ &  $93,043_{8588}$ & $0.51_{0.016}$ & $0.189_{0.0115}$ & $187.07_{4.650}$ & $1.82_{0.012}$ & $-0.0279_{0.00321}$ \\
8  &  $2,093_{70}$ & $214,844_{6793}$ & $0.41_{0.007}$ & $0.098_{0.0038}$ & $205.29_{2.615}$ & $1.92_{0.006}$ & $-0.0263_{0.00184}$ \\
10 &  $3,619_{61}$ & $348,761_{5275}$ & $0.33_{0.004}$ & $0.053_{0.0011}$ & $192.76_{1.150}$ & $2.05_{0.009}$ & $-0.0199_{0.00127}$ \\
12 &  $5,657_{59}$ & $476,614_{3416}$ & $0.27_{0.002}$ & $0.030_{0.0005}$ & $168.50_{1.003}$ & $2.29_{0.012}$ & $-0.0141_{0.00072}$ \\
14 &  $8,352_{60}$ & $594,902_{2459}$ & $0.23_{0.001}$ & $0.017_{0.0002}$ & $142.46_{0.652}$ & $2.56_{0.007}$ & $-0.0102_{0.00044}$ \\
16 & $11,797_{63}$ & $707,326_{2296}$ & $0.21_{0.001}$ & $0.010_{0.0001}$ & $119.92_{0.368}$ & $2.72_{0.003}$ & $-0.0080_{0.00036}$ \\
17 & $13,795_{77}$ & $762,197_{2299}$ & $0.20_{0.001}$ & $0.008_{0.0001}$ & $110.51_{0.377}$ & $2.79_{0.005}$ & $-0.0072_{0.00026}$ \\
\hline
\end{tabular}
\end{center}
\end{table*}

Notice that the mean number of vertexes ($\bar n_v$) confirms that
the number of local optima (and thus the search difficulty)
increases with the value of $K$. Some other interesting inferences
can be drawn from these metrics. First of all, looking at the $\bar
l$ values one can conclude that the maxima networks are small worlds
for all values of $K$ since the growth of $\bar l$ is bounded by a
function $O(log \; n_v)$. In a sense, this is not surprising as the
whole configuration space spans the binary hypercube $\{0,1\}^N$ of
degree $N$ with $2^N$ vertices, which has maximum distance
(diameter) $d=log 2^N$, i.e. 16 and 18 for our studied instances.
However, while the base configuration space has constant degree for
any node, the maxima network are degree-inhomogeneous (see next
section) and have clustering coefficients well above those of
equivalent random graphs, showing that there is local structure in
the networks. For both $N = 16$ and 18, the mean degree $\bar z$
first increases with $K$ and then goes down again for $K>8$. The
assortativity coefficients are always very small which means that
there is almost no correlation between the degrees of neighboring
nodes. For easy energy landscapes, Doye found that the networks were
slightly disassortative ~\cite{doye05}.

\subsection{Degree Distributions}

\begin{figure*} [!ht]
\vspace{-0.1cm}
\begin{center}
\begin{tabular}{cc}
    \mbox{\includegraphics[width=5.5cm]{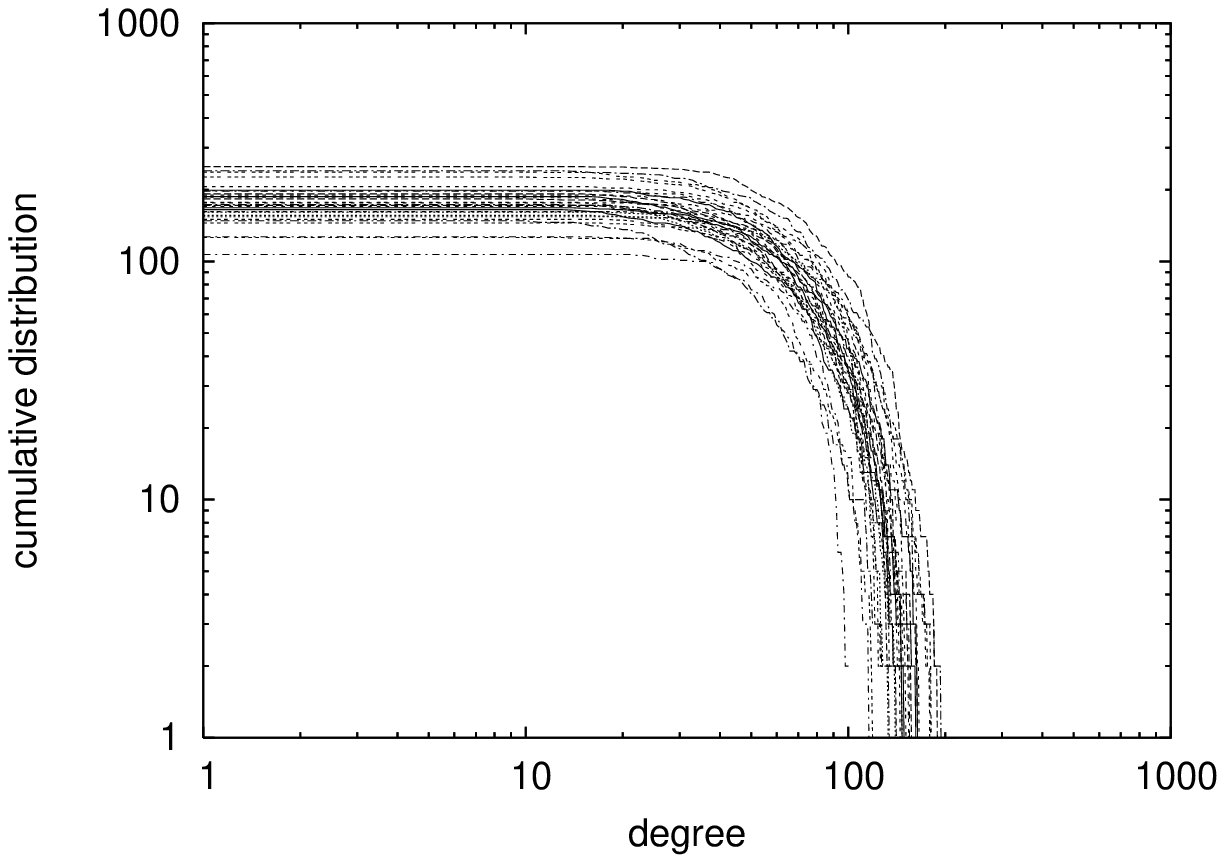} } \protect &
    \mbox{\includegraphics[width=5.5cm]{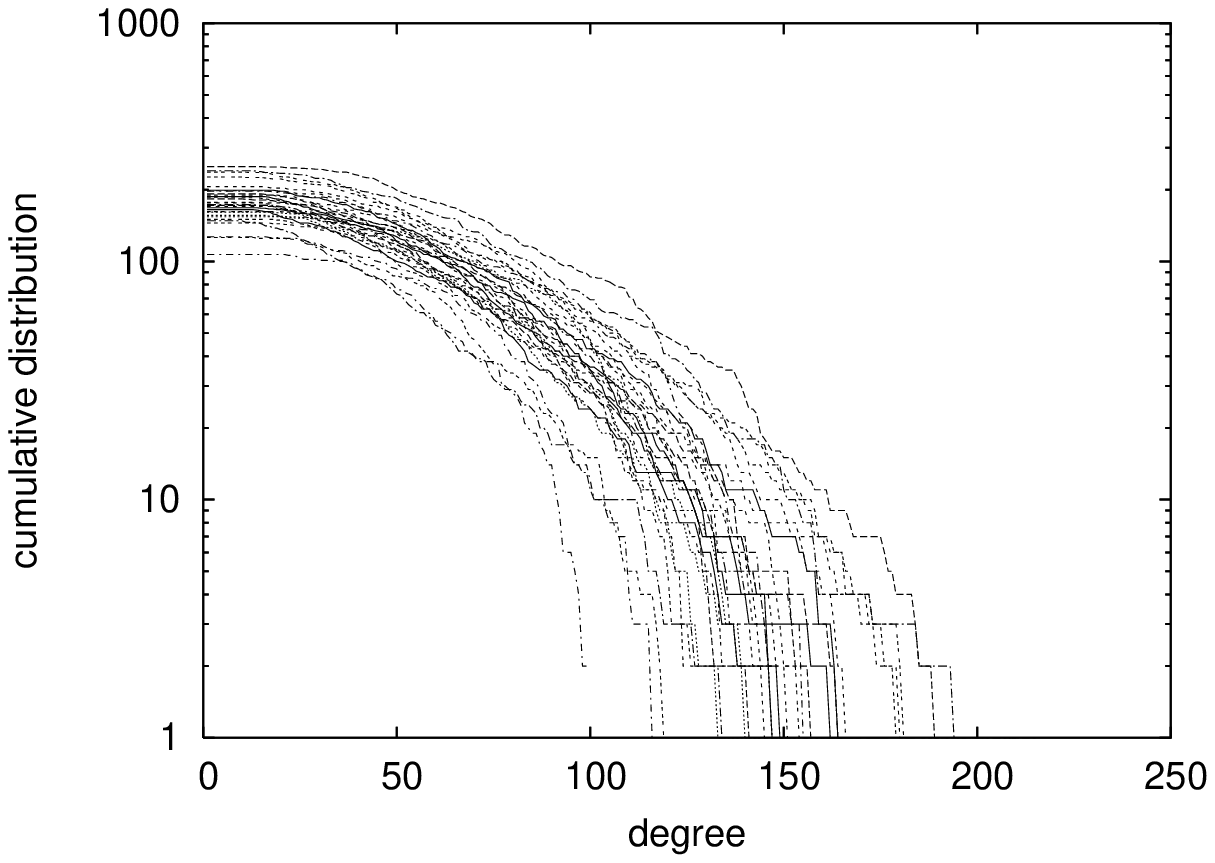} } \protect \\
    \mbox{\includegraphics[width=5.5cm]{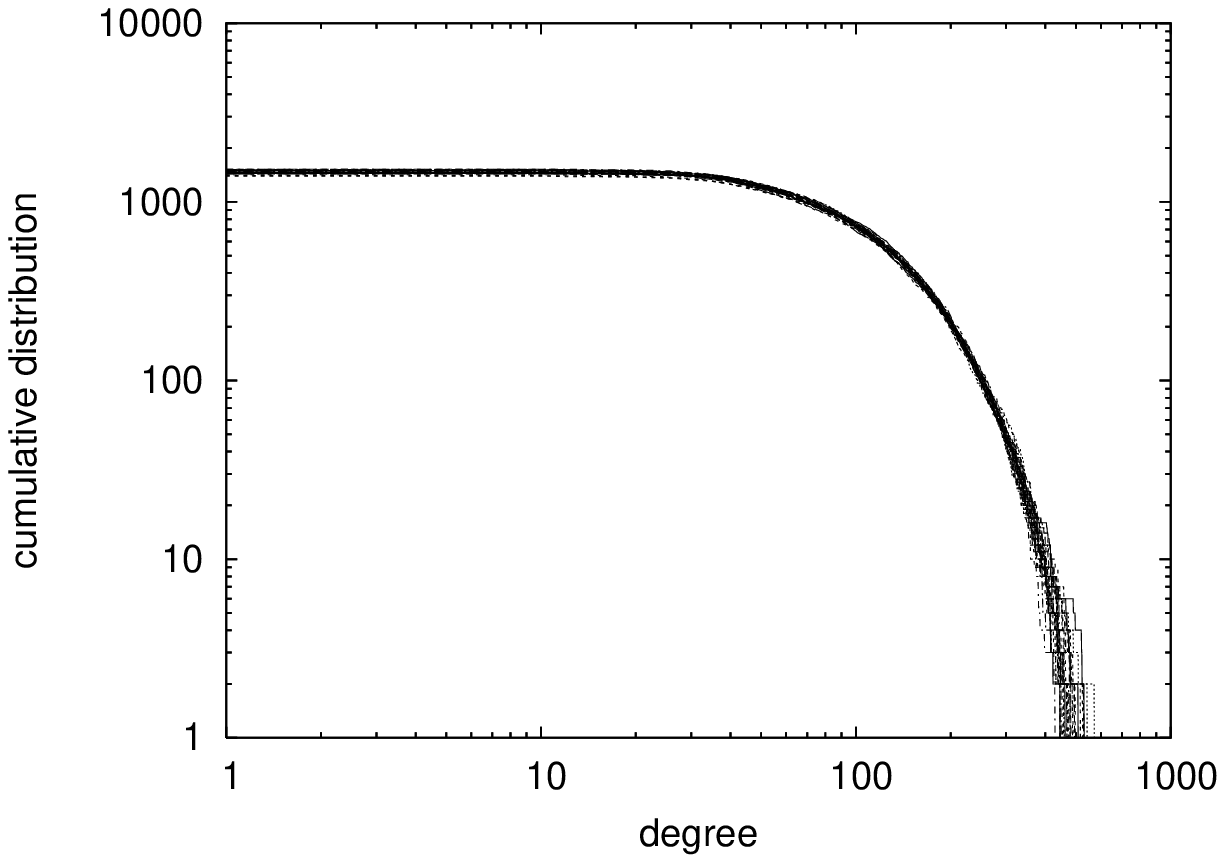} } \protect &
    \mbox{\includegraphics[width=5.5cm]{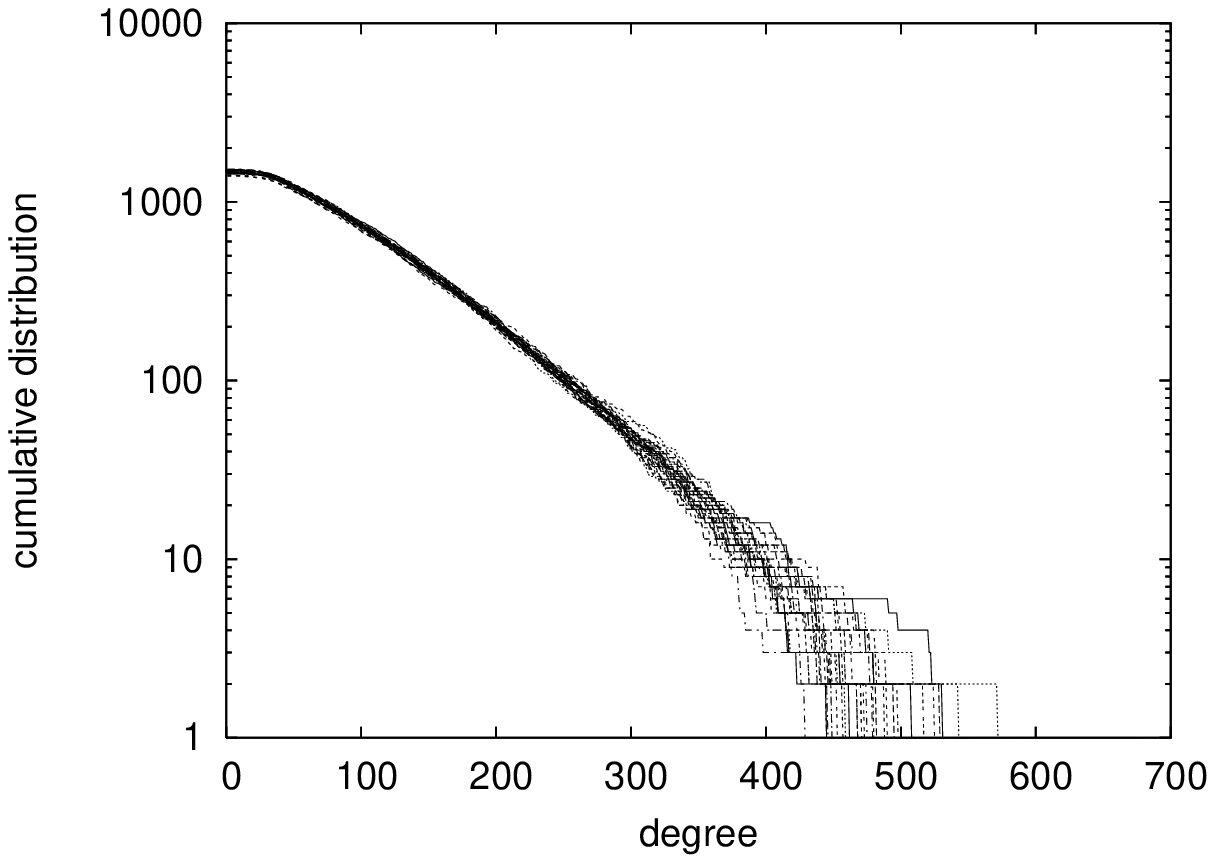} } \protect \\
    \end{tabular}
\caption{Cumulative degree distributions for $N=16$ and $K=4$ (top),
$K=10$ (bottom). All 30 curves are plotted. Left: Log-log plot.
Right: Lin-log plot.\label{DFN16K4K10}}
\end{center}
\end{figure*}

\begin{figure*} [!ht]
\begin{center}
\begin{tabular}{cc}
    \mbox{\includegraphics[width=5.5cm]{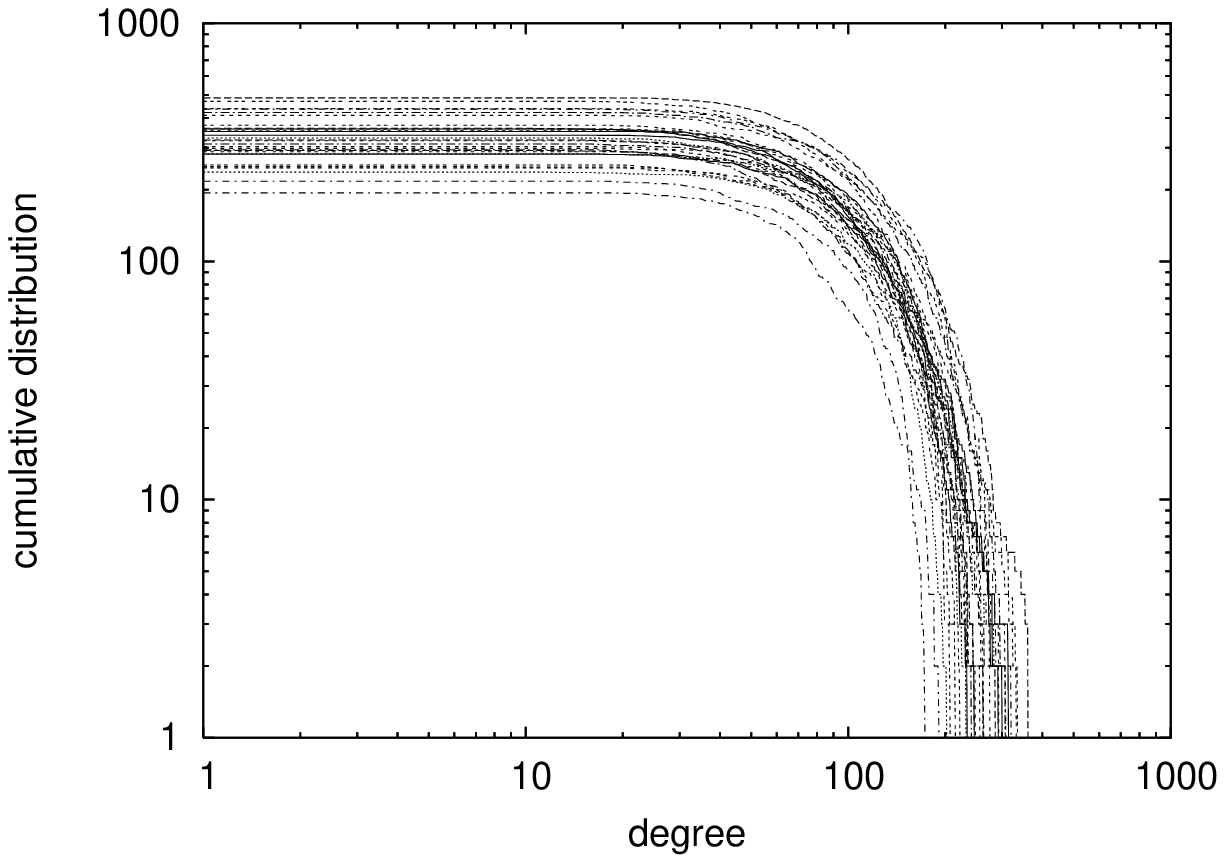} } \protect &
    \mbox{\includegraphics[width=5.5cm]{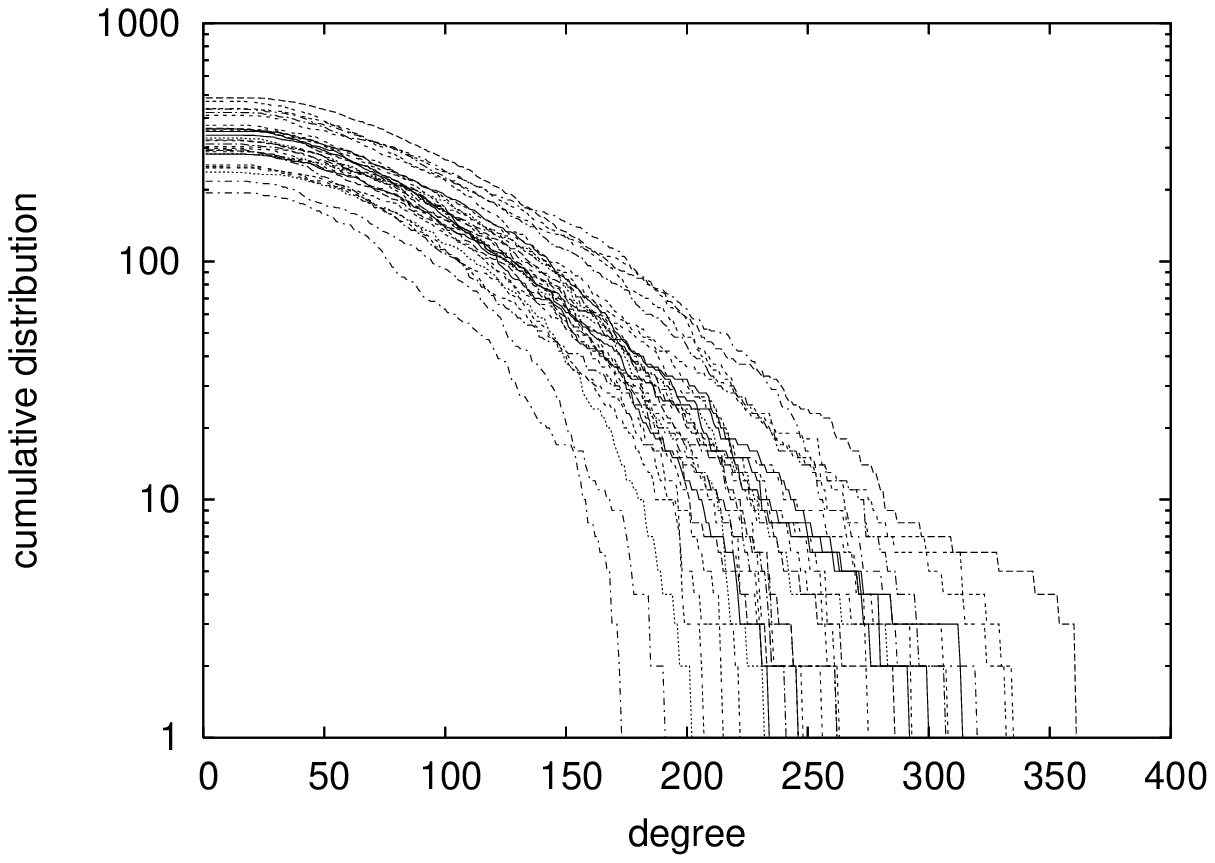} } \protect \\
    \mbox{\includegraphics[width=5.5cm]{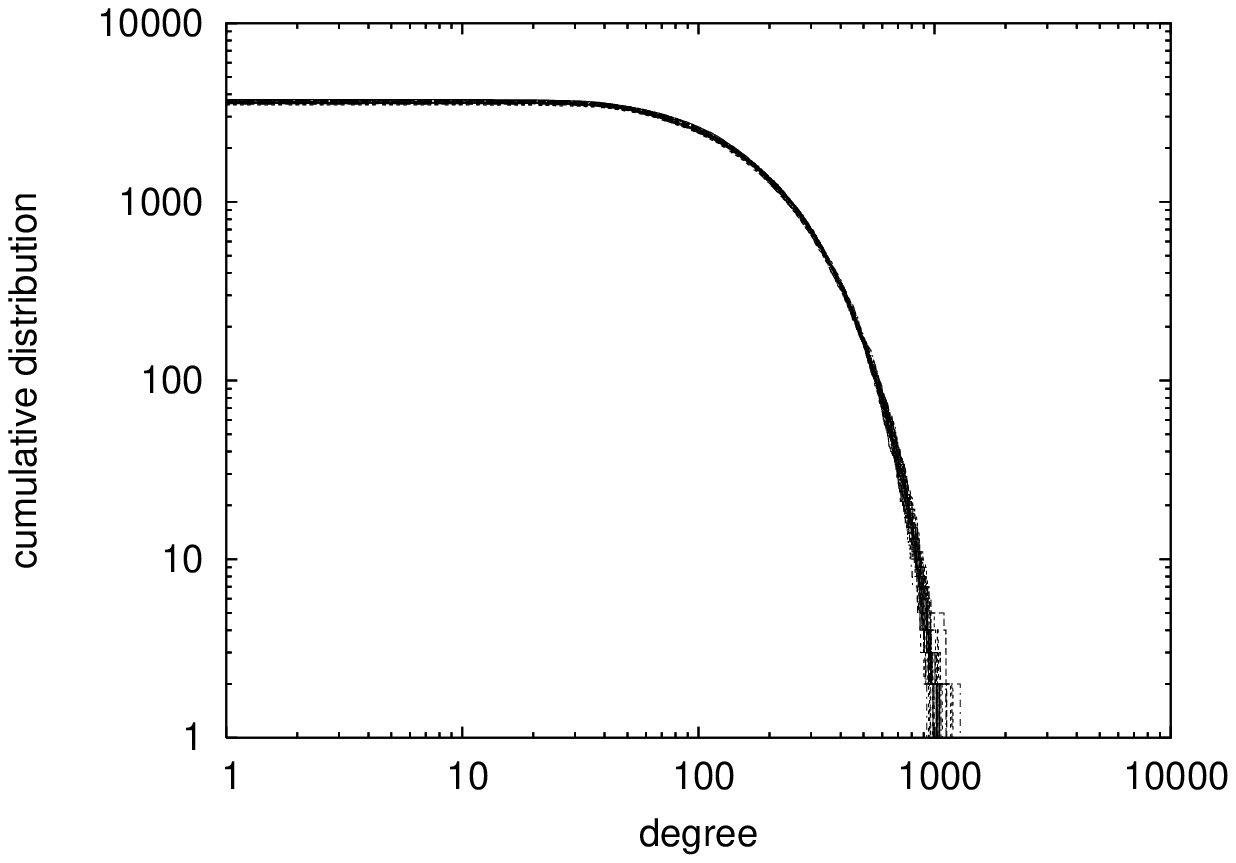} } \protect &
    \mbox{\includegraphics[width=5.5cm]{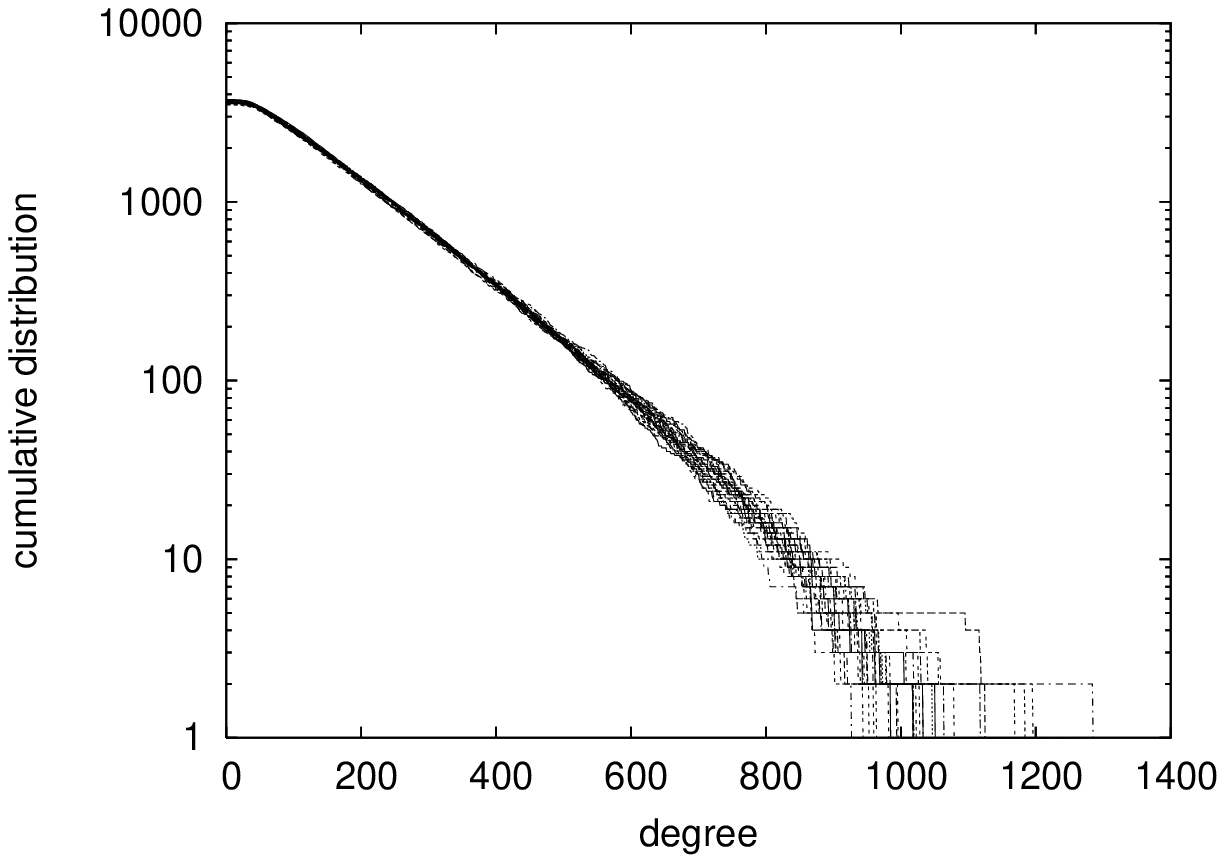} }  \protect \\
    \end{tabular}
\caption{Cumulative degree distributions for $N=18$ and $K=4$ (top),
$K=10$ (bottom). All 30 curves are plotted. Left: Log-log plot.
Right: Lin-log plot.\label{DFN18K4K10}}
\end{center}
\end{figure*}

The degree distribution function $p(k)$ of a  graph represents
the probability that a randomly chosen node has degree
$k$~\cite{newman03}. Random graphs are characterized by a $p(k)$ of
Poissonian form, while social and technological real networks often
show long tails to the right, i.e. there are nodes that have an
unusually large number of neighbors. Sometimes this behavior can be
described by a power-law, but often the distribution is less extreme
and can be fitted by a stretched exponential or by an exponentially
truncated power-law\cite{newman03}.

Figure~\ref{DFN16K4K10} shows all the curves for 30 randomly
generated landscapes for $N = 16$ and $K=4,10$, whilst figure
~\ref{DFN18K4K10} does the same    for $N = 18$. To smooth out
fluctuations in the high degree region, the cumulative degree
distribution function is plotted, which is just the probability that
the degree is greater than or equal to $k$. The single curves are
shown rather than the average curve because the sum of a sufficient
number of independent random variables with arbitrary distributions,
provided that the first few moments exist and are finite, tends to
distribute normally according to a general formulation of the
central limit theorem~\cite{Feller}. In other words, if the average
of the sum were plotted, the original shapes would essentially be
lost. The curves cannot be described by power-laws: this possibility
is ruled out by the left parts of figs.~\ref{DFN16K4K10},
and~\ref{DFN18K4K10} which are double logarithmic plots. In log-log
plots, power laws should appear as straight lines, at least for a
sizable part of abscissae range.

\begin{figure} [!ht]
\begin{center}
\includegraphics[width=5.5cm] {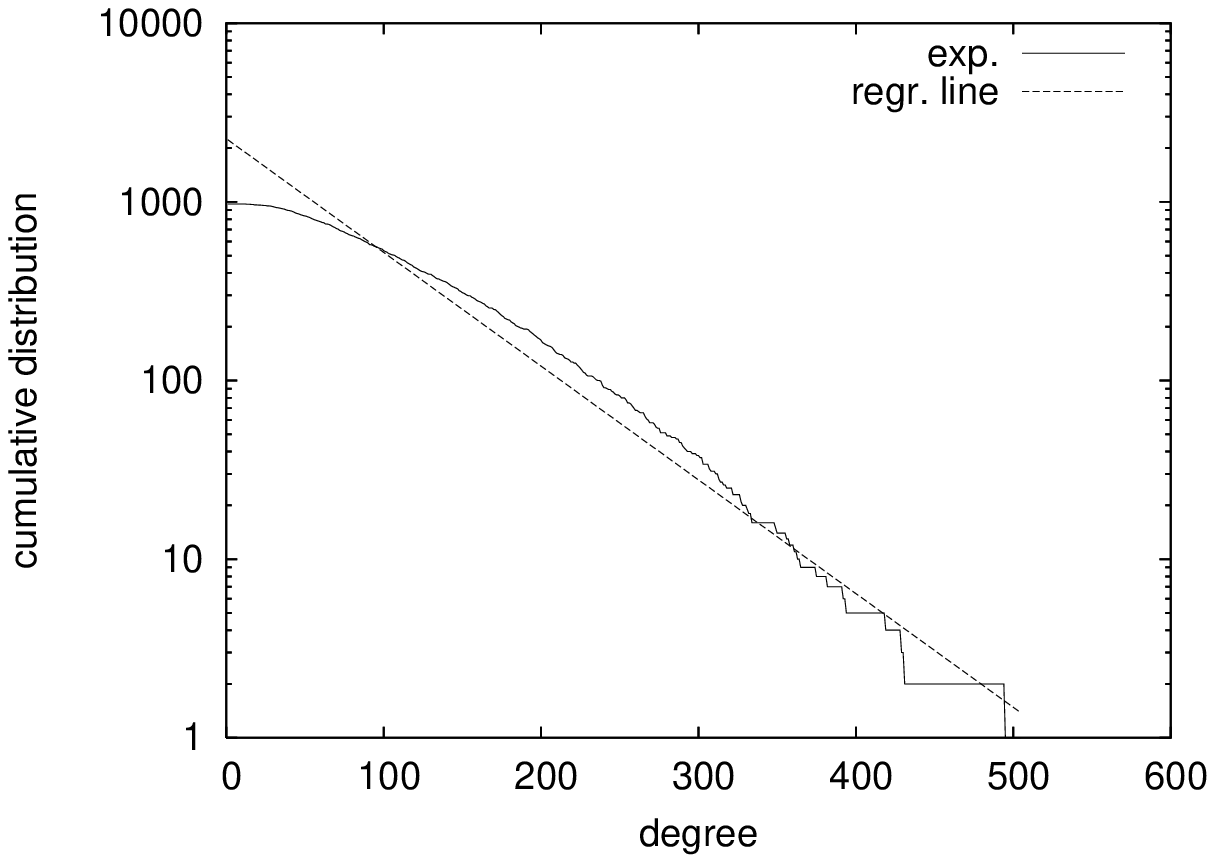} \protect \\
\includegraphics[width=5.5cm] {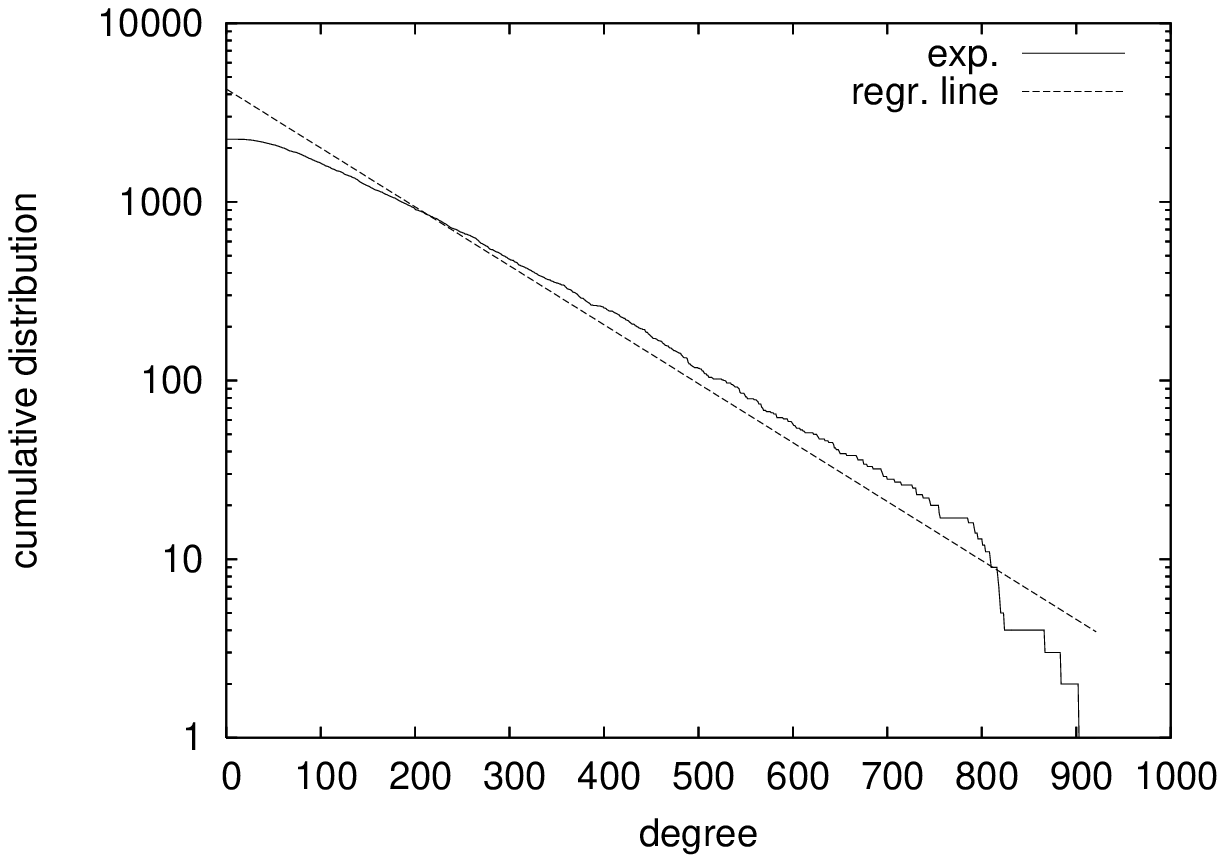} \protect \\

\caption{Cumulative degree distribution (with regression line) of
two representative instances with $K=8$, $N=16$ (top) and $N=18$
(bottom).} \label{fig:dloglog}
\end{center}
\end{figure}

On the other hand, the right images in the same figures show that
the distributions can be fitted approximately by exponentials of the
type $p(k)=(1/z) e^{-k/z}$ where $z$ is the mean degree, as most
curves are approximately straight lines on these linear-log plots.
This is true for the larger part of the degree range. When we
approach the finite degree cutoff the fit is obviously less good. Small
networks such as those with $N=16$ and $K=4$  show larger
fluctuations and their tails decay faster than exponentially. Two particular examples
with a medium value of $K$ ($K=8$) are shown in detail in fig.~\ref{fig:dloglog}, together
with an exponential fit. Table~\ref{tab:cumulDeg} gives the parameters of the regression
lines for all $N$ and $K$ values.

\begin{table}[!ht]
\begin{center}
\small \caption{Correlation coefficient ($\bar{\rho}$), intercept
($\bar{\alpha})$ and slope ($\bar{\beta}$) and slope of the linear
regression between the cumulative number of nodes and the degree of
nodes : $\log(p(k)) = \alpha + \beta k + \epsilon$. The averages and
standard deviations of 30 independent landscapes, are shown.}
\label{tab:cumulDeg}
\begin{tabular}{|c|c|c|c|}
\hline
\multicolumn{4}{|c|}{$N = 16$} \\
\hline
 $K$ & $\bar \rho$ & $\bar \alpha$ & $\bar{\beta}$ \\
\hline
2  & $-0.816_{0.340}$ & $4.05_{0.717}$ & $-0.1109_{0.0379}$ \\
4  & $-0.932_{0.026}$ & $6.07_{0.276}$ & $-0.0295_{0.0026}$ \\
6  & $-0.967_{0.009}$ & $7.09_{0.105}$ & $-0.0178_{0.0009}$ \\
8  & $-0.986_{0.006}$ & $7.60_{0.107}$ & $-0.0144_{0.0007}$ \\
10 & $-0.989_{0.004}$ & $8.04_{0.125}$ & $-0.0146_{0.0008}$ \\
12 & $-0.990_{0.004}$ & $8.51_{0.156}$ & $-0.0170_{0.0010}$ \\
14 & $-0.992_{0.003}$ & $8.92_{0.121}$ & $-0.0202_{0.0010}$ \\
15 & $-0.991_{0.004}$ & $9.11_{0.144}$ & $-0.0220_{0.0011}$ \\
\hline
\hline
\multicolumn{4}{|c|}{$N = 18$} \\
\hline
2  & $-0.823_{0.343}$ & $4.57_{0.865}$  & $-0.1088_{0.0325}$  \\
4  & $-0.951_{0.025}$ & $6.71_{0.225}$  & $-0.0198_{0.0021}$  \\
6  & $-0.982_{0.007}$ & $7.74_{0.107}$  & $-0.0098_{0.0005}$  \\
8  & $-0.991_{0.004}$ & $8.28_{0.096}$  & $-0.0076_{0.0003}$  \\
10 & $-0.994_{0.003}$ & $8.74_{0.119}$  & $-0.0076_{0.0004}$  \\
12 & $-0.995_{0.003}$ & $9.19_{0.161}$  & $-0.0088_{0.0005}$  \\
14 & $-0.995_{0.003}$ & $9.65_{0.134}$  & $-0.0110_{0.0005}$  \\
16 & $-0.994_{0.003}$ & $10.1_{0.173}$  & $-0.0139_{0.0008}$  \\
17 & $-0.994_{0.005}$ & $10.2_{0.207}$  & $-0.0151_{0.0008}$  \\
\hline
\end{tabular}
\end{center}
\end{table}

If we compare these results with Doye's~\cite{doye02,doye05} the
most important difference is that we do not observe power-law
distributions. Indeed, power-law degree distributions of the
inherent energy landscape networks point to the ``easiness'' of
those landscapes: due to the presence of highly connected nodes,
which are also among the fittest, a simple gradient-descent would
bring a searcher down to a local energy minimum, often the global
one, starting anywhere in the configuration space. In other words,
there exist the ``funnel'' effect described by Doye~\cite{doye02}.
In contrast, $NK$ landscapes have tunable difficulty. How can random
networks with exponential degree distributions be obtained? One way
is the following: in each time step, just add a new node, and add a
new link between two randomly chosen nodes, including the new one.
Iterating this dynamical process produces graphs with an exponential
distribution of the node degrees~\cite{Dorogovtsev-Mendes-03}. But
$NK$ landscapes are static and thus it is difficult to see how this
process could be implemented. However, the following qualitative
explanation might help. Imagine that $K$ is increased from 2 to
$N-1$ in single steps. Then we could have the image of the previous
landscape increasing its size and deforming itself when $K$ goes
from its current value to $K+1$. The new maxima that appear could be
considered as if they were added dynamically (of course some
previous optima might disappear as well). Edges in the new landscape
are selected essentially randomly, with more probability of
selecting an already existing node. Thus, with this imaginary
mechanism a distribution close to exponential would be obtained.

Thus, as the $NK$ landscape difficulty varies smoothly when $K$ is
increased, the degree distribution of the corresponding maxima
networks remains essentially exponential. We do not observe
scale-free distributions for the easy landscapes as in the energy
landscape case~\cite{doye02}. This is understandable: standard
energy landscapes in molecular chemistry and crystal physics do
correspond to thermodynamically stable states which are naturally
smooth and easy to reach when the system is forming or it is
slightly perturbed. In contrast, $NK$ landscape are synthetic and do
not correspond to any physical principle in their construction. The
only physical systems that resemble $NK$ landscapes are spin
glasses, in which conflicting energy minimization requirements lead
to frustration and to landscape ruggedness~\cite{kauffman93,Stein}.
However,  disordered condensed matter systems similar to spin
glasses are only obtained in particular situations, for instance by
fast cooling~\cite{Stein}.

\subsection{Basins of Attraction}
\label{basins}

Besides the maxima network, it is useful to describe the associated
basins of attraction as these play a key role in search algorithms.
Furthermore, some characteristics of the basins can be related to
the network features described above. The notion of the basin of
attraction of a local maximum has been presented in
sect.~\ref{defs}. We have exhaustively computed the size and number
of all the basins of attraction for $N=16$ and $N=18$ and for all
even $K$ values plus $K=N-1$. In this section, we analyze the basins
of attraction from several points of view as it is described below.

\subsubsection{Global optimum basin size vs. $K$}

\begin{figure} [!ht]
\begin{center}
\includegraphics[width=7cm] {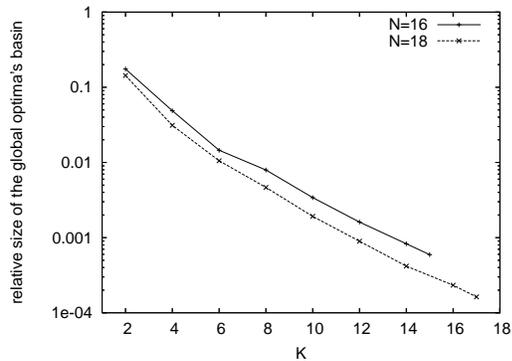} \protect \\
\caption{Average of the relative size of the basin corresponding to the global
maximum for each K over 30 landscapes.\label{nbasins}}
\end{center}
\end{figure}

In Figure~\ref{nbasins} we plot the average size of the basin
corresponding to the global maximum for $N=16$ and $N=18$, and all
values of $K$ studied. The trend is clear: the basin shrinks very
quickly with increasing $K$. This confirms that the higher the $K$
value, the more difficult for an stochastic search algorithm to
locate the basin of attraction of the global optimum

\subsubsection{Number of basins of a given size}

\begin{figure} [!ht]
\begin{center}
    \mbox{\includegraphics[width=6cm]{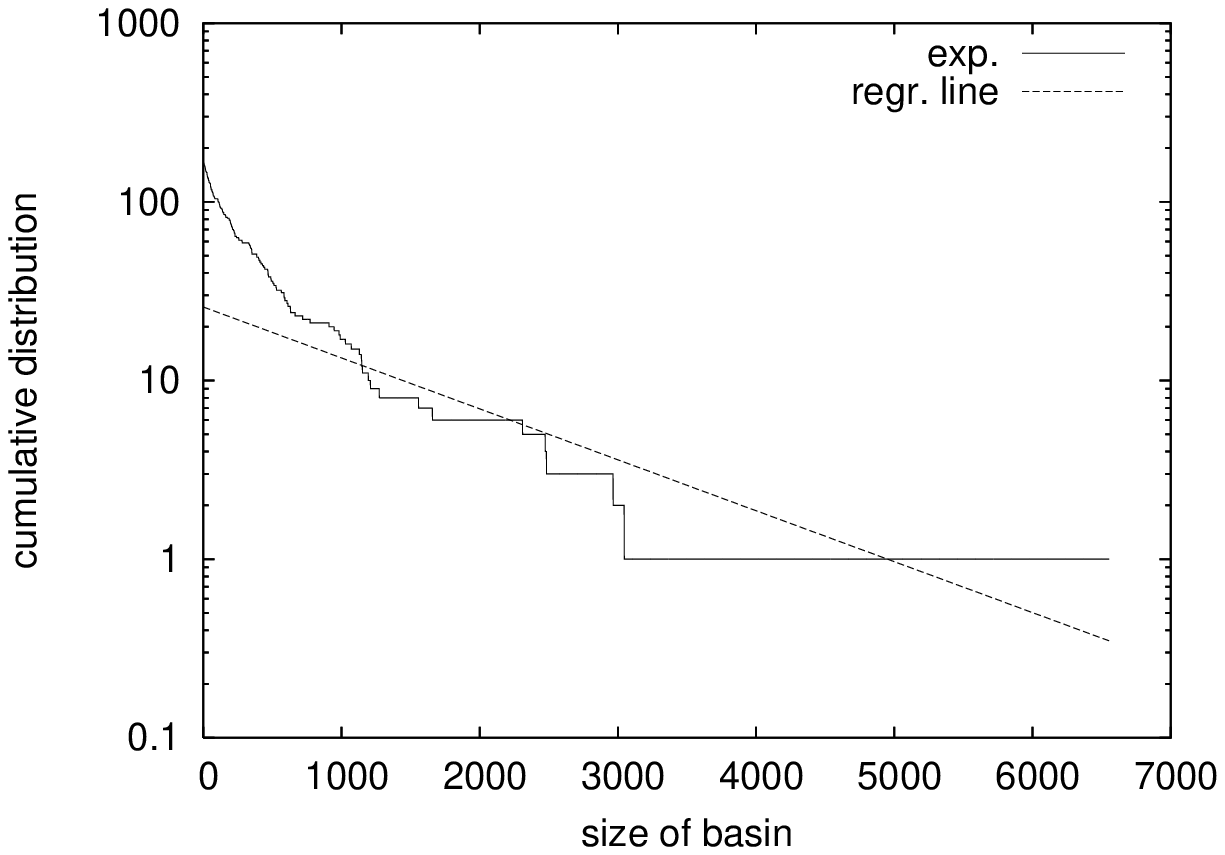} } \protect \\
    \mbox{\includegraphics[width=6cm]{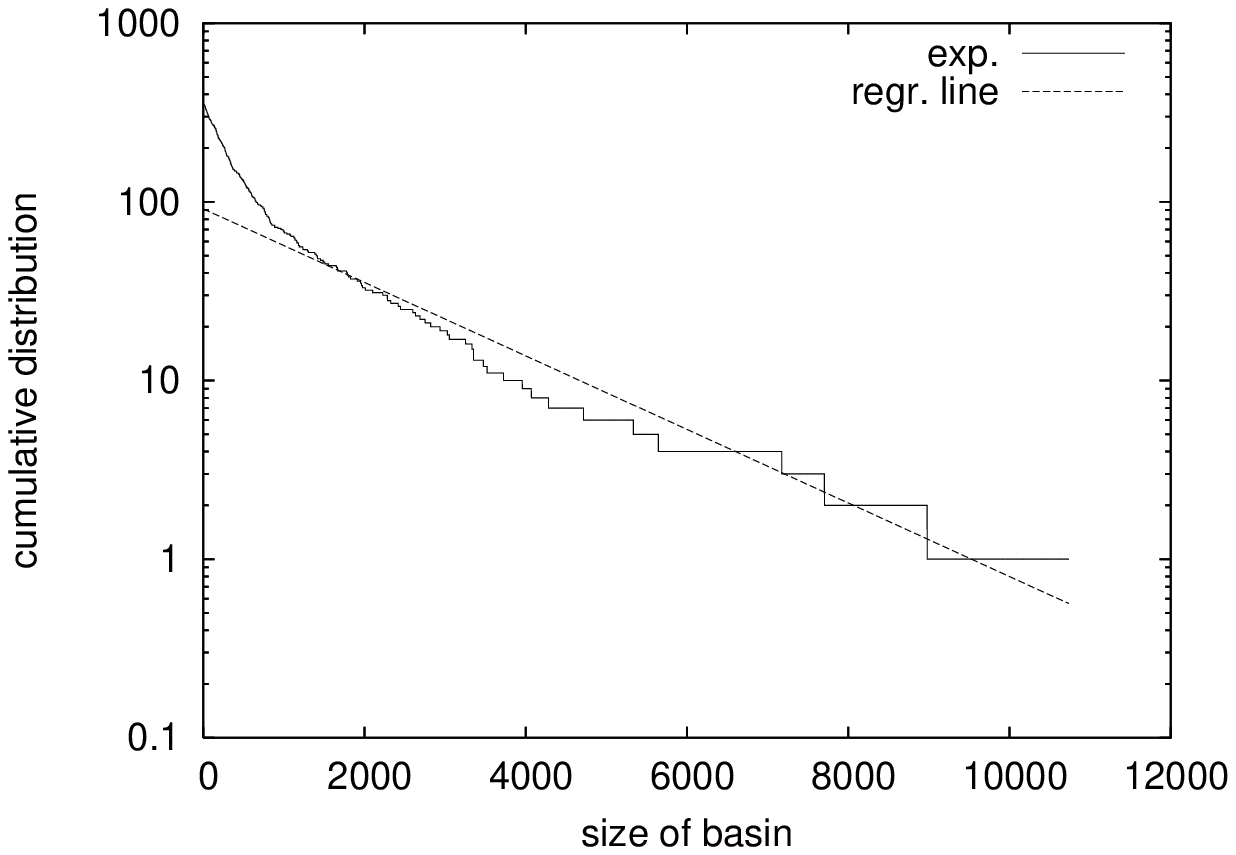} } \protect \\
\caption{Cumulative distribution of the number of basins of a given size with regression line.
Two Representative landscapes are visualized with N=16 (top) and N=18 (bottom) and K=4. A lin-log scale is used. \label{bas-18-size}}
\end{center}
\end{figure}

\begin{table}[!ht]
\begin{center}
\small \caption{Correlation coefficient ($\bar{\rho}$), and linear
regression coefficients (intercept ($\bar{\alpha})$ and slope
($\bar{\beta}$)) of the relationship between the basin size of
optima and the cumulative number of nodes of a given (basin) size (
in logarithmic scale: $\log(p(s)) = \alpha + \beta s + \epsilon$).
The average and standard deviation values over 30 instances, are
shown.}

\label{tab:cumulSize} \vspace{0.2cm}
\begin{tabular}{|c|c|c|c|}
\hline
\multicolumn{4}{|c|}{$N = 16$} \\
\hline
 $K$ & $\bar \rho$ & $\bar \alpha$ & $\bar{\beta}$ \\
\hline
2   & $-0.944_{0.0454}$ & $2.89_{0.673}$ & $-0.0003_{0.0002}$  \\
4   & $-0.959_{0.0310}$ & $4.19_{0.554}$ & $-0.0014_{0.0006}$  \\
6   & $-0.967_{0.0280}$ & $5.09_{0.504}$ & $-0.0036_{0.0010}$  \\
8   & $-0.982_{0.0116}$ & $5.97_{0.321}$ & $-0.0080_{0.0013}$  \\
10  & $-0.985_{0.0161}$ & $6.74_{0.392}$ & $-0.0163_{0.0025}$  \\
12  & $-0.990_{0.0088}$ & $7.47_{0.346}$ & $-0.0304_{0.0042}$  \\
14  & $-0.994_{0.0059}$ & $8.08_{0.241}$ & $-0.0508_{0.0048}$  \\
15  & $-0.995_{0.0044}$ & $8.37_{0.240}$ & $-0.0635_{0.0058}$  \\
\hline
\hline
\multicolumn{4}{|c|}{$N = 18$} \\
\hline
2   & $-0.959_{0.0257}$ & $3.18_{0.696}$ & $-0.0001_{0.0001}$  \\
4   & $-0.960_{0.0409}$ & $4.57_{0.617}$ & $-0.0005_{0.0002}$  \\
6   & $-0.967_{0.0283}$ & $5.50_{0.520}$ & $-0.0015_{0.0004}$  \\
8   & $-0.977_{0.0238}$ & $6.44_{0.485}$ & $-0.0037_{0.0007}$  \\
10  & $-0.985_{0.0141}$ & $7.24_{0.372}$ & $-0.0077_{0.0011}$  \\
12  & $-0.989_{0.0129}$ & $7.98_{0.370}$ & $-0.0150_{0.0019}$  \\
14  & $-0.993_{0.0072}$ & $8.69_{0.276}$ & $-0.0272_{0.0024}$  \\
16  & $-0.995_{0.0056}$ & $9.33_{0.249}$ & $-0.0450_{0.0036}$  \\
17  & $-0.992_{0.0113}$ & $9.49_{0.386}$ & $-0.0544_{0.0058}$  \\
\hline
\end{tabular}
\end{center}
\end{table}

Figure~\ref{bas-18-size} shows the cumulative distribution of the
number of basins of a given size (with regression line) for two
representative instances with $K = 4$ and $N=16$ (top) and $N=18$.
Table ~\ref{tab:cumulSize} shows the average (of 30 independent
landscapes) correlation coefficients and linear regression
coefficients (intercept ($\bar{\alpha})$ and slope ($\bar{\beta}$))
between the number of nodes and the basin sizes. Notice that
distribution decays exponentially or faster for the lower $K$ and it
is closer to exponential for the higher $K$. This observation is
relevant to theoretical studies that estimate the size of attraction
basins (see for example \cite{garnier01}). These studies  often
assume that the basin sizes are uniformly distributed. From the
slopes $\bar \beta$ of the regression lines
(table~\ref{tab:cumulSize}) one can see that high values of $K$ give
rise to steeper distributions (higher $\bar \beta$ values). This
indicates that there are less basins of large size for large values
of $K$. In consequence, basins are broader for low values of $K$,
which is consistent with the fact that those landscapes are
smoother.

\subsubsection{Fitness of local optima vs. their basin sizes}

\begin{figure} [!ht]
\begin{center}
    \mbox{\includegraphics[width=6cm]{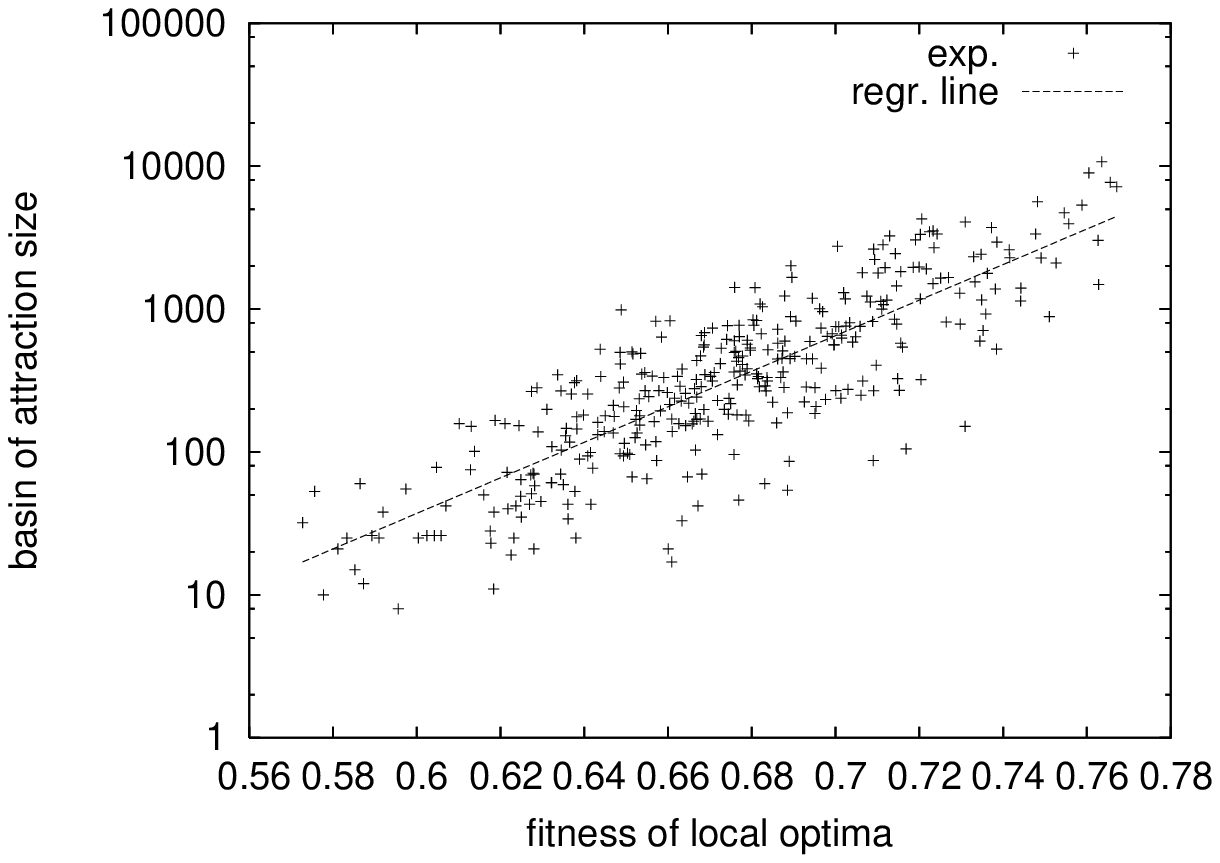} } \protect \\
    \mbox{\includegraphics[width=6cm]{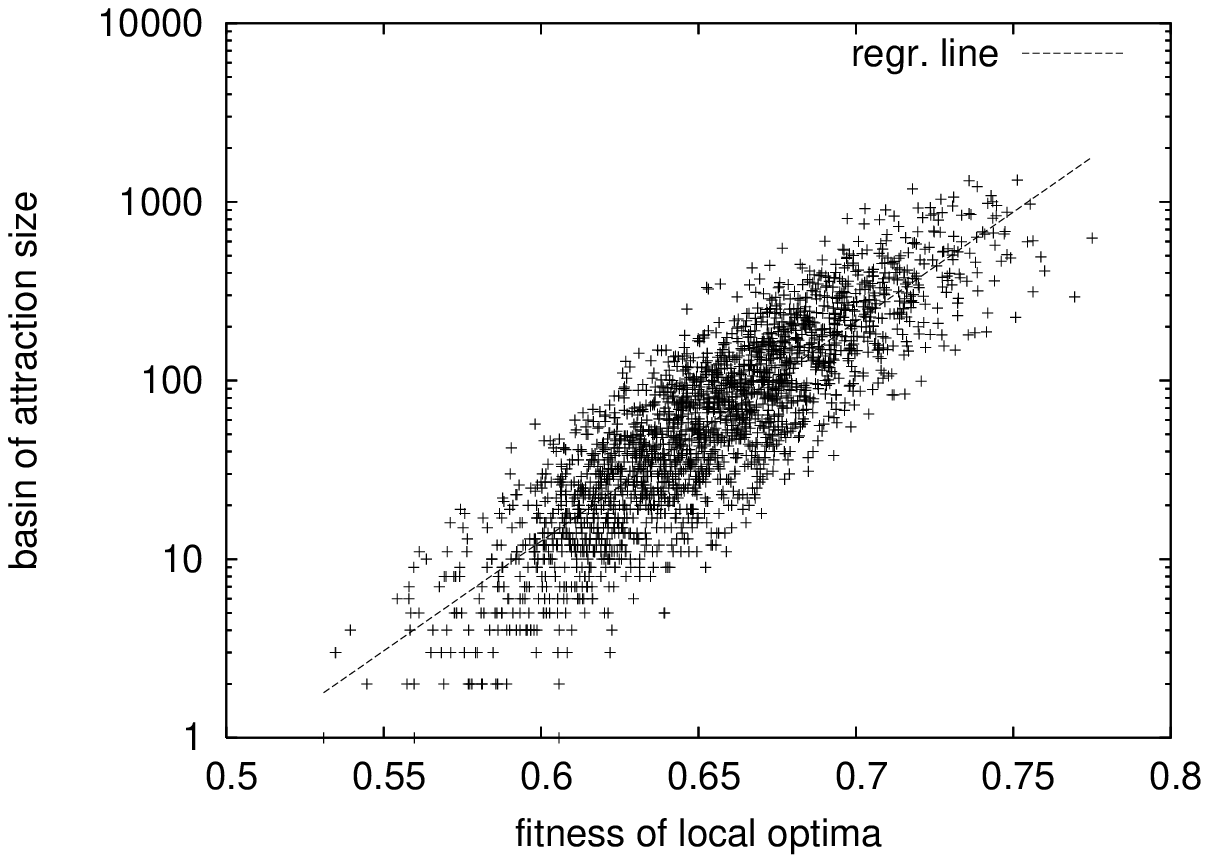} } \protect \\
\caption{Correlation between the fitness of local optima and their corresponding basin sizes, for two representative
instances with $N=18$, $K=4$ (top) and $K=8$ (bottom). \label{fig:cor_fit-size}}
\end{center}
\end{figure}

\begin{table}[!ht]
\begin{center}
\small \caption{Correlation coefficient ($\bar{\rho}$), and linear
regression coefficients (intercept ($\bar{\alpha})$ and slope
($\bar{\beta}$)) of the relationship between the fitness of optima
and their basin size (in logarithmic scale: $\log(s) = \alpha +
\beta f + \epsilon$). The average and standard deviation values over
30 instances, are shown} \label{tab:cor_fit-size} \vspace{0.2cm}
\begin{tabular}{|c|c|c|c|}
\hline
\multicolumn{4}{|c|}{$N = 16$} \\
\hline
 $K$ & $\bar \rho$ & $\bar \alpha$ & $\bar{\beta}$ \\
\hline
2   & $0.832_{0.0879}$ & $-15.476_{5.9401}$ & $33.066_{8.9252}$  \\
4   & $0.842_{0.0259}$ & $-13.035_{1.9907}$ & $27.094_{2.8611}$  \\
6   & $0.852_{0.0180}$ & $-12.977_{0.9921}$ & $26.061_{1.4908}$  \\
8   & $0.860_{0.0088}$ & $-12.570_{0.3769}$ & $24.880_{0.5725}$  \\
10  & $0.850_{0.0050}$ & $-11.954_{0.3501}$ & $23.561_{0.5421}$  \\
12  & $0.833_{0.0065}$ & $-11.485_{0.2993}$ & $22.519_{0.4773}$  \\
14  & $0.816_{0.0047}$ & $-11.261_{0.2008}$ & $21.864_{0.3256}$  \\
15  & $0.812_{0.0044}$ & $-11.352_{0.2109}$ & $21.876_{0.3298}$  \\
\hline
\hline
\multicolumn{4}{|c|}{$N = 18$} \\
\hline
2   & $0.839_{0.0680}$ & $-16.585_{6.0606}$ & $35.925_{8.6640}$  \\
4   & $0.842_{0.0257}$ & $-14.458_{2.1746}$ & $30.174_{3.1520}$  \\
6   & $0.852_{0.0140}$ & $-14.542_{0.9596}$ & $29.219_{1.4147}$  \\
8   & $0.867_{0.0066}$ & $-14.515_{0.3750}$ & $28.538_{0.5988}$  \\
10  & $0.866_{0.0038}$ & $-13.914_{0.3068}$ & $27.209_{0.4621}$  \\
12  & $0.854_{0.0030}$ & $-13.180_{0.1700}$ & $25.751_{0.2804}$  \\
14  & $0.836_{0.0027}$ & $-12.602_{0.1399}$ & $24.553_{0.2214}$  \\
16  & $0.822_{0.0022}$ & $-12.502_{0.1039}$ & $24.133_{0.1633}$  \\
17  & $0.817_{0.0027}$ & $-12.583_{0.1278}$ & $24.143_{0.2066}$  \\
\hline
\end{tabular}
\end{center}
\end{table}

The scatter-plots in figure~\ref{fig:cor_fit-size} illustrate the
correlation between the basin sizes of local maxima (in logarithmic
scale) and their fitness values. Two representative instances for
$N$ = 18 and $K$ = 4,8 are shown. Table ~\ref{tab:cor_fit-size}
shows the averages (of 30 independent landscapes) of the correlation
coefficient, and the linear regression coefficients between these
two metrics (maxima fitness and their basin sizes). All the studied
landscapes for $N$ = 16 and 18, are reported. Notice that, there is
a clear positive correlation between the fitness values of maxima
and their basins' sizes. In other words, the higher the peak the
wider tend to be its basin of attraction. Therefore, on average,
with a hill-climbing algorithm, the global optimum would be easier
to find than any other local optimum. This may seem surprising. But,
we have to keep in mind  that as the number of local optima
increases (with increasing $K$), the global optimum basin is more
difficult to reach by an stochastic local search algorithm (see
figure~\ref{nbasins}). This observation offers a mental picture of
$NK$ landscapes: we can consider the landscape as composed of a
large number of mountains (each corresponding to a basin of
attraction), and those mountains are wider the taller the hilltops.
Moreover, the size of a mountain basin grows exponentially with its
hight.

\subsubsection{Basins sizes of local optima vs. their degrees}

\begin{figure} [!ht]
\begin{center}
    \mbox{\includegraphics[width=6cm]{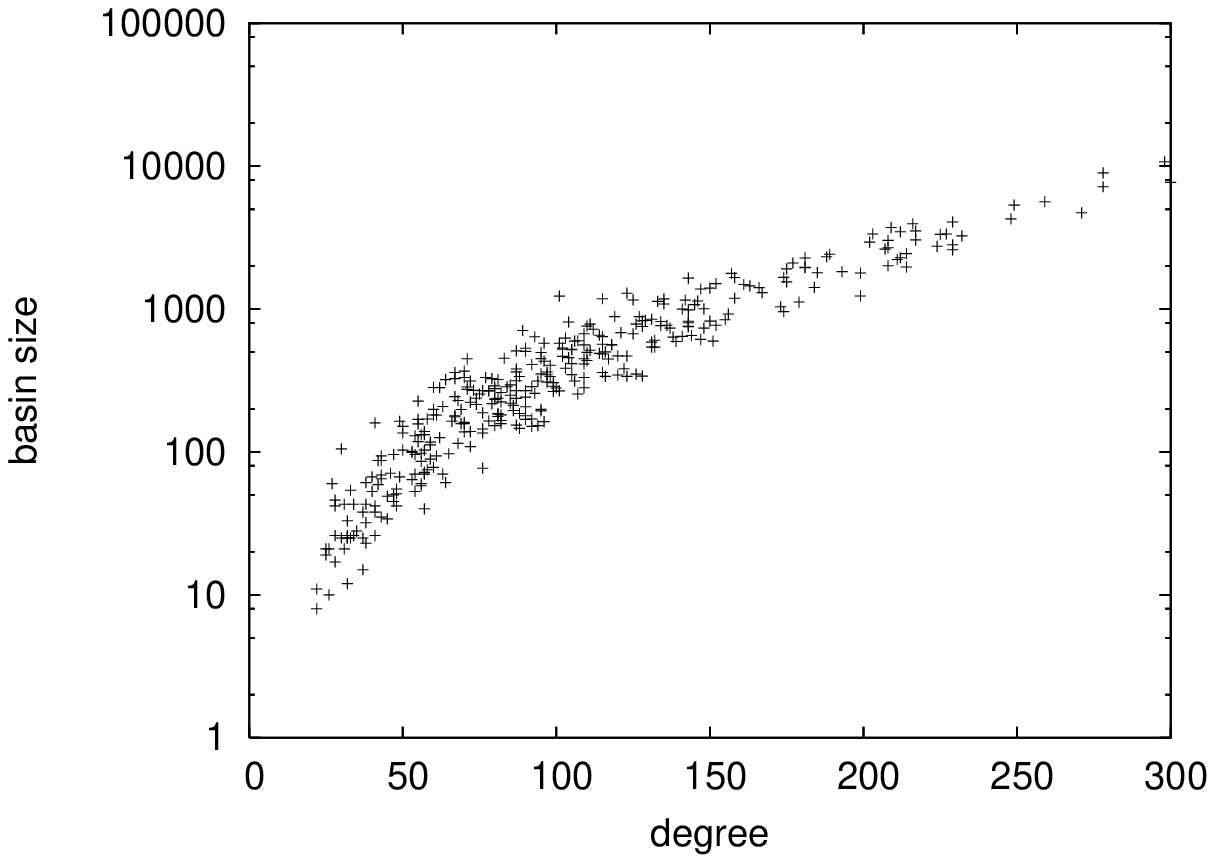} } \protect \\
    \mbox{\includegraphics[width=6cm]{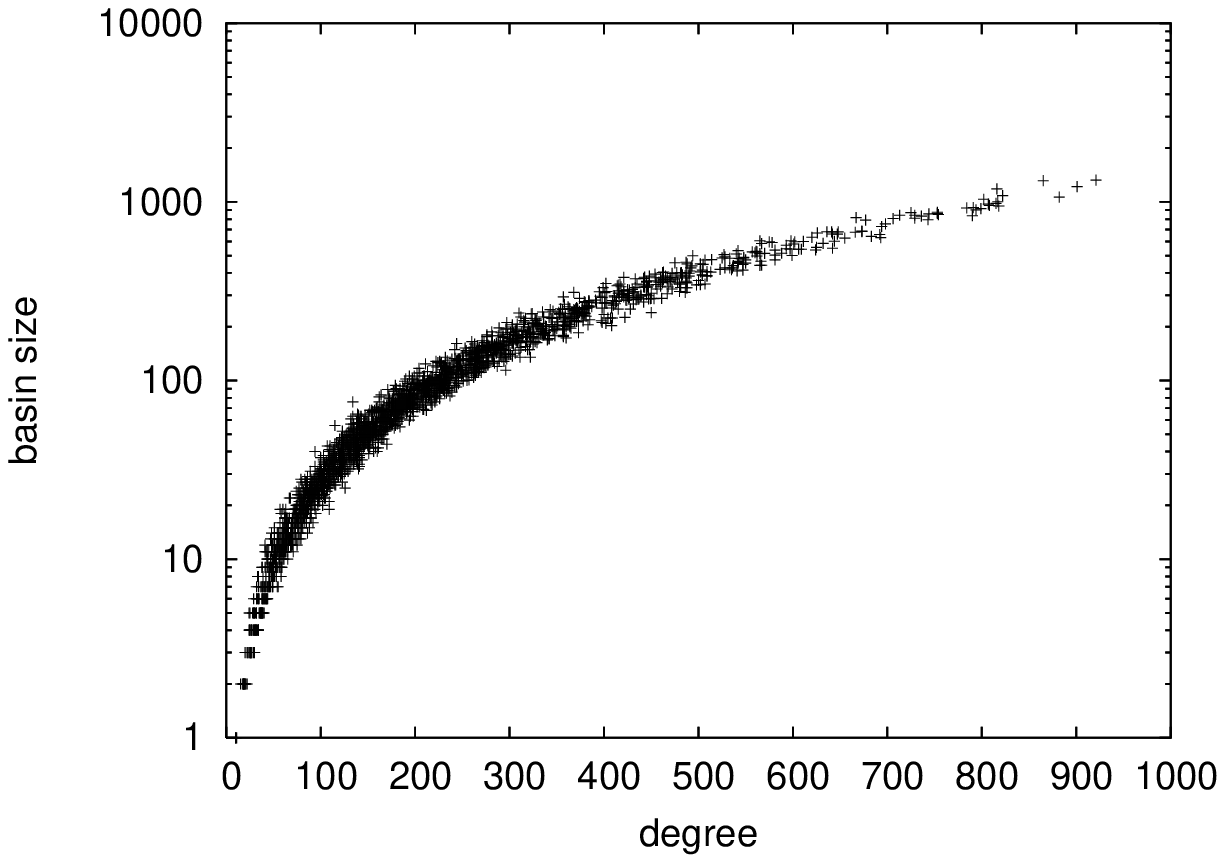} } \protect \\
\caption{Correlation between the degree of local optima and their corresponding basin sizes, for two representative instances
with with $N=18$, $K=4$ (top) and $K=8$ (bottom). \label{fig:degree-bsize1}}
\end{center}
\end{figure}

The scatter plots in figure~\ref{fig:degree-bsize1} illustrate the
correlation between basin sizes of maxima and their degrees.
Representative instances with $N=18$, and $K=4, 8$, are illustrated.
There is a clear positive correlation between the degree and the
basin sizes of maxima in the network. This observation suggests that
landscapes with low $K$ values can be searched more effectively
since a given configuration has many neighbors belonging to the same
large basin of attraction. It is also confirmed that the basins for
low $K$ are much larger than those for high $K$, not only the basin
corresponding to the global maximum.

\section{Conclusions}

We have proposed a new characterization of combinatorial fitness
landscapes using the well-known family of $NK$ landscapes as an
example. We have used an extension of the concept of inherent
networks proposed for energy surfaces~\cite{doye02} in order to
abstract and simplify the landscape description. In our case the
inherent network is the graph where the vertices are all the local
maxima and edges mean basin adjacency between two maxima. We have
exhaustively obtained these graphs for $N=16$ and $N=18$, and for
all even values of $K$, plus $K=N-1$. The maxima graphs are small
worlds since the average path lengths are short and scale
logarithmically in the size of the graphs. However, the maxima
graphs are not random. This is shown by their clustering
coefficients, which are much larger than those of corresponding
random graphs and also   by their degree distribution
functions, which are not Poissonian but rather exponential. The
construction of the maxima networks requires the determination of
the basins of attraction of the corresponding landscapes. We have
thus described the nature of the basins and their relationship with
the local maxima network. We have found  that the size of the basin
corresponding to the global maximum becomes smaller with increasing
$K$. The distribution of the basin sizes is approximately
exponential for all $N$ and $K$, but the basin sizes are larger for
low $K$, another indirect indication of the increasing randomness
and difficulty of the landscapes when $K$ becomes large. Finally,
there is a strong positive correlation between the basin size of a
maxima and their degree, which confirms that the synthetic view
provided by the maxima graph is a useful one.

This study represents our first attempt towards a topological and
statistical characterization of easy and hard combinatorial
landscapes. Much remains to be done. First of all, the results found
should be confirmed for larger instances of $NK$ landscapes. This
will require good sampling techniques, or theoretical studies since
exhaustive sampling becomes quickly impractical. Other landscape
types should also be examined, such as those containing neutrality,
which are very common in real-world applications. Work is in
progress for neutral versions of $NK$ landscapes. Finally, the
landscape statistical characterization is only a step toward
implementing good methods for searching it. We thus hope that our
results will help in designing or estimating efficient search
techniques and operators.

\label{conclusions}

\providecommand{\bysame}{\leavevmode\hbox
to3em{\hrulefill}\thinspace}
\providecommand{\MR}{\relax\ifhmode\unskip\space\fi MR }
\providecommand{\MRhref}[2]{%
  \href{http://www.ams.org/mathscinet-getitem?mr=#1}{#2}
} \providecommand{\href}[2]{#2}

\balancecolumns
\end{document}